\documentclass[10pt,twocolumn,letterpaper]{article}

\usepackage[]{cvpr}      % To produce the REVIEW version
\usepackage{multirow}
\definecolor{cvprblue}{rgb}{0.21,0.49,0.74}
\usepackage[pagebackref,breaklinks,colorlinks,allcolors=cvprblue]{hyperref}

\def\paperID{*****} % *** Enter the Paper ID here
\def\confName{CVPR}
\def\confYear{2026}
\newcommand{\ours}{TokTalk}

\title{\ours: Expressive Real-time Facial Animation from Audio-LLM Tokens}

\author{Qingcheng Zhao \quad Yifang Pan \quad Karan Singh\\
University of Toronto\\
Toronto, Canada\\
{\tt\small qc.zhao@mail.utoronto.ca \quad evan.pan@mail.utoronto.ca \quad karan@dgp.toronto.edu}
}

\begin{document}

\twocolumn[{
\renewcommand\twocolumn[1][]{#1}
\maketitle
\centering
\includegraphics[width=\textwidth]{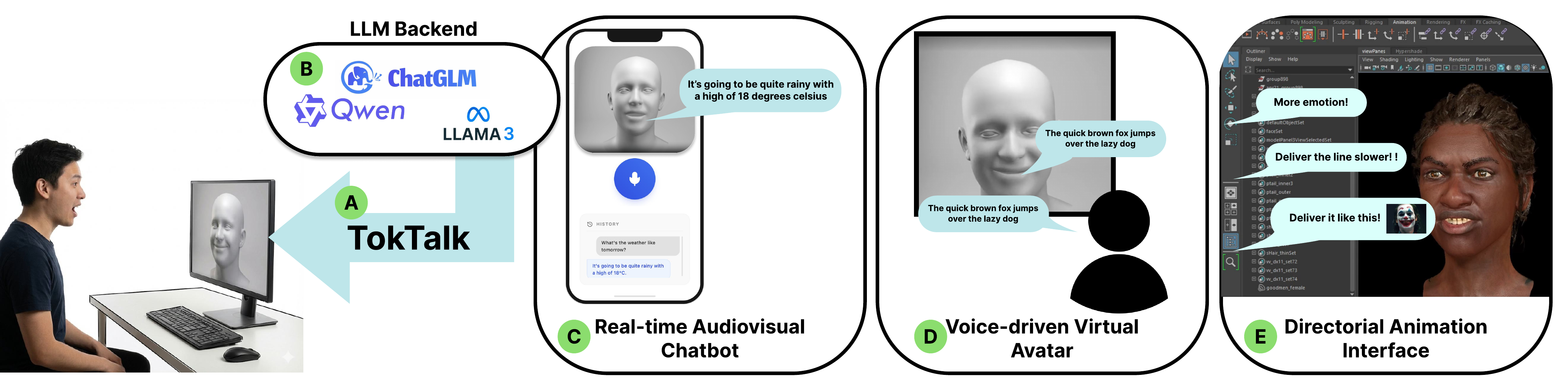}
\captionof{figure}{TokTalk is an audio-token based facial animation system for expressive, low-latency audiovisual applications. Our system (A) produces expressive facial motion in parallel with audio, directly from audio-tokens of state-of-the-art audio LLMs through adaptation (B). TokTalk enables a range of real-time audiovisual 3D face applications: expressive chatbots (C), voice-driven virtual avatars by processing real-time voice input through audio tokenizers (D), and a voice driven animation Director's interface that showcases our ability to generate and control a diverse range audio-visual facial performances (E).}
\label{fig:teaser}
}]

\begin{abstract}
Recent advances in Audio-LLM's like GPT-4o have ushered in an era of conversational interaction with language models.
Conversational avatars however, still seem robotic in facial expression and conversational flow, in part due to sequential stages of speech recognition, text generation, turn-based text response, speech synthesis, and audio driven facial animation. Based on our insight that audio-tokens produced by current Audio-LLMs carry sufficient information to reconstruct a plausible facial performance, we present {\em \ours}, a system that directly outputs expressive facial animation in real-time from streaming audio-tokens.
We construct a novel audio-token to 3D facial motion dataset, on which {\em TokTalk} is trained using a Chunk-based Conditional Flow Matching model. A lightweight adaptation strategy allows our trained model to seamlessly connect to any token-based Audio-LLM at minimal computational overhead. Our chunk-based processing further enables parametric trade-off between latency and facial quality, shown through ablation studies.
We further show that the real-time performance of {\em TokTalk} is comparable in latency to prior art solutions, and significantly favorable (via a perceptual study) in terms of quality, expressivity and control of the 3D facial performance. We showcase {\em TokTalk}'s flexibility using  
a chatbot avatar, a voice-driven user avatar, and an animation director's interface, as diverse audio-visual face applications.  
\end{abstract}

\section{Introduction}
\label{sec:intro}
Natural face-to-face conversations between humans and digital bots, once the domain of science fiction like in the film Her \cite{her2013}, are now tantalizingly close to reality. The advent of audio-LLM's like GPT-4o have shown the profound potential of conversational interaction with large language models.
Conversational avatars however, still seem robotic in facial expression and conversational flow, in part due to the cascaded pipeline by which they are constructed: A user speaks; the speech in turn is turned to text; to which a textual LLM response is generated; the response text is synthesized into speech audio; and this audio in turn re-encoded to audio-features used to map to a 3D animated face. The sequential stages introduce unnecessary latency, and expressive audio-visual information can get lost across stages of encoding and decoding depending on their design objectives. {\em TokTalk} addresses this problem.

\begin{figure}[h]
    \centering
    \includegraphics[width=1\linewidth]{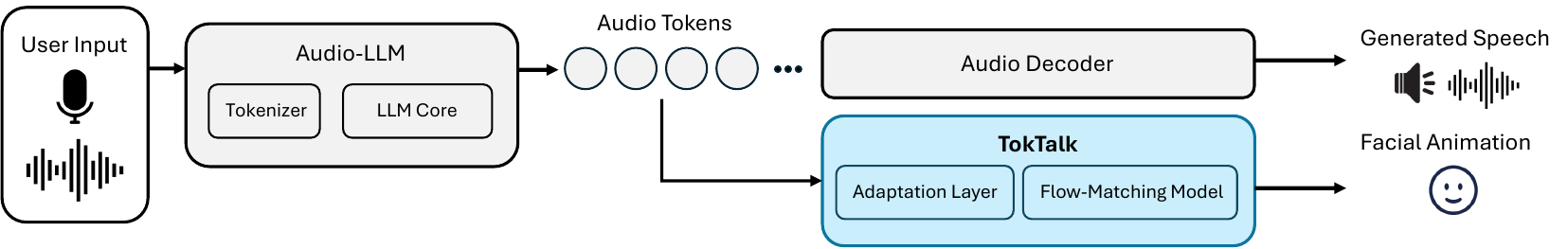}
    \caption{{\em TokTalk} pipeline: existing audio-LLM (gray); our face module (blue).}
    \label{fig:voice_llm}
\end{figure}

We observe that pre-trained speech encoders like Wav2Vec~2.0 \cite{baevski2020wav2vec}, and HuBERT \cite{hsu2021hubert}, which are used by prior work to
generate 3D facial performance, were primarily optimized for automatic speech recognition (ASR), aligning audio signals with their textual content. Such optimization largely discards a wealth of non-textual information crucial for expressive facial animation such as: hesitant pauses, sighs, stutters, laughter, breaths, or fearful/angry tremors in a voice. This makes both emotion and paralingual motion \cite{cyberpunk2020} difficult to reconstruct on the 3D face.
We show that in contrast, the internal tokens embedding of modern generative Audio-LLMs, designed for high-fidelity speech audio reconstruction and semantic understanding, carry rich audio information pertaining to non-verbal and emotional nuance, that ASR-based features ignore
(Table~\ref{tab:ravdess_emotion}).
{\em TokTalk} exploits this insight by using these audio tokens, to directly generate plausible and expressive 3D facial motion in real-time (Fig.~\ref{fig:voice_llm}).

Prior art in real-time speech-driven facial animation roughly fall into three categories. Procedural approaches like JALI ~\cite{edwards_jali_2016} that combine audio features and an aligned speech transcript, produce high-quality animator editable output, but are constrained by the latency of synthesized audio features and sufficient phonetic context, that impedes their use in expressive real-time applications.
Deterministic data-driven methods like the CNN-based \cite{Tiny_sig25} afford low latency but have limited control over expression due to limited model capacity and deterministic mapping.
While recent 2D diffusion models like Ditto ~\cite{li2025ditto} and MuseTalk ~\cite{zhang2025musetalk} demonstrate the feasibility of high-quality real-time synthesis, 3D diffusion-based methods like ~\cite{zhao_media2face_2024, sun_diffposetalk_2023, pan2025modelb} suffer from unacceptably slow inference for real-time use. 
Our flow-based model bridges this gap, delivering diffusion-based expressiveness and real-time performance.

{\bf Overview}:\\
Our method takes audio-tokens from an Audio-LLM as input, and generates vivid, synchronized facial animations in parallel with audio generation, reducing the cascading latency significantly (Fig.~\ref{fig:voice_llm}). 
We first construct a novel token-to-facial-animation dataset using an audio-LLM tokenizer .  
Our system is designed as a lightweight "plug-in" module of an Audio-LLM system. 
We train a simple yet effective Conditional Flow Matching Model that maps the audio tokens directly to FLAME parameters \cite{FLAME:SiggraphAsia2017}. This "LLM-aligned" approach enables the facial animation to be generated in parallel with the audio generation stream, eliminating the additional latency incurred by the traditional first-audio-then-animation framework. Another advantage of our plug-in architecture is its adaptability:
our model can easily adapt to work with any other audio-LLM via a lightweight adaptation layer. Our chunk-based processing further enables parametric trade-off between latency and facial quality, shown through our ablation studies.

\begin{table*}[h]
    \centering
    \caption{Comparison of audio driven facial animation approaches based on their ability to provide: paralingual motion (eg. conversational brow and head motion); synchronized control over editing facial expression or vocal delivery; real-time performance.}
    \label{tab:method_comparison}
    \begin{tabular}{lcccc}
        \hline
        \hline
        \textbf{Method} & \textbf{Paralingual} & \textbf{\begin{tabular}[c]{@{}c@{}}Face \\ control\end{tabular}} & \textbf{\begin{tabular}[c]{@{}c@{}}Voice \\ control\end{tabular}} & \textbf{Real-time} \\
        \hline
        \citet{peng2023emotalk}        & \checkmark & \checkmark & $\times$ & $\times$ \\
        \citet{FaceDiffuser_Stan_MIG2023} & \checkmark & $\times$   & $\times$ & $\times$ \\
        \citet{sun_diffposetalk_2023}      & \checkmark & \checkmark & $\times$ & $\times$ \\
        \citet{Tiny_sig25}               & $\times$   & $\times$   & $\times$ & \checkmark \\
        \citet{lee2025audio}             & \checkmark & \checkmark & $\times$ & \checkmark \\
        \citet{edwards_jali_2016}        & \checkmark & \checkmark & $\times$ & $\times$ \\
        \hline
        \textit{TokTalk} (Ours)          & \checkmark & \checkmark & \checkmark & \checkmark \\
        \hline
        \hline
    \end{tabular}
\end{table*}

{\bf Evaluation:}\\
{\em TokTalk} embodies the design goals of a modular approach to expressive 3D facial animation in real-time, that animates both vocal and paralingual aspects of the face \cite{cyberpunk2020}, parallelizes and decouples control over the audio-visual creation and editing of an entire facial performance, while keeping them perfectly synchronized. While prior approaches share some of the features of our work, none of them address all our design goals (Table~\ref{tab:method_comparison}).
We show that the real-time performance of {\em TokTalk} is comparable in latency to prior art solutions, and significantly favorable (via a perceptual study) in terms of quality, expressivity and control of the 3D facial performance. 
We showcase {\em TokTalk}'s flexibility using three diverse proof-of-concept applications: a chatbot avatar for use with Audio-LLMs, a voice-driven user avatar that could be used in immersive games and other VR applications, and an animation Director's interface, to conversationally direct an autonomous digital actor.

{\bf Contributions:}\\
Our primary contribution is exploiting the insight that Audio-LLM tokens can be used to directly drive the motion of a 3D face, as shown by {\em TokTalk} (the first such system to our knowledge). {\em TokTalk} further contributes:

\begin{itemize}
    \item Validation that audio-LLM tokens carry richer spoken, emotional, and paralingual information than ASR discriminative audio features.
    \item A parallel approach to producing facial animation concurrently with audio (as the latency bottleneck).
    \item A modular adaptation framework enabling compatibility across diverse Audio-LLM architectures without re-training our flow-matching model.
\end{itemize}

\section{Related Work}
\label{sec:related}
This section surveys the landscape of audio-driven lipsync animation, audio representations, and audio-LLMs.

\subsection{Audio-driven 3D Facial Animation}
Early research in facial animation primarily focused on lip synchronization and can be broadly categorized into two groups: procedural and data-driven.\\
\textbf{Procedural methods} like \cite{pan_vocal_2022, edwards_jali_2016, bailly_animated_2012, cohen_modeling_1993}, are based on the insight that a face can be plausibly animated to produce speech by mapping phonemes (sounds) to corresponding visual visemes (mouth shapes). Such approaches can lack fine detail and facial nuance but are well suited to animator control. \\
\textbf{Data-driven methods} learn the correlation between audio representations and 3D motion using an \textbf{audio encoder} and an \textbf{audio-conditioned motion decoder} \cite{xing_codetalker_2023, richard_meshtalk_2022, faceformer2022, karras_audio-driven_2017, jung2024speed, liu2025medtalk}. These data-driven approaches shows high quality results with fine detail but produce facial motion that is hard to edit and control.

Contemporary models extend beyond the mouth to paralingual motion of the entire face. Since the correlation between audio and paralingual expression is weaker, these methods often necessitate the use of additional conditioning signals to ensure the performance fits the spoken context. Works such as Emotalk \cite{peng2023emotalk} and EMOTE \cite{danvevcek2023emotional} use categorical emotional labels to control the affect of the delivery. More recently, diffusion models have enabled sophisticated forms of conditioning: Media2face \cite{zhao_media2face_2024} utilizes text embeddings as a primary control signal, while Diffposetalk \cite{sun_diffposetalk_2023} and \cite{pan2025modelb} leverage reference motion style to generate animations with fine-grained control. Despite their expressive power, the iterative nature of diffusion models introduces significant inference latency, precluding their use in real-time applications.

To meet the demands of conversational chatbots and digital humans—such as Grok's Ani \cite{ani} and Soul Machines \cite{soul}, recent research has shifted toward minimizing generation latency. \citet{lee2025audio} addresses this by utilizing a causal audio encoder to reduce encoding delay and employing Distribution Matching Distillation to derive a high-speed diffusion model capable of generating motion in 4 diffusion steps, achieving a per-frame runtime of 10ms. \citet{Tiny_sig25} pushes this strategy further by jointly distilling both the audio encoder and the motion decoder, reaching runtimes as low as 2ms per frame. However, because these models require fully processed audio as input, their latency gain comes from distilling the model, which limits the model's expressiveness \cite{Tiny_sig25} and introduces jitter artifacts \citet{lee2025audio}. Our design in contrast, alleviates sequential latency by moving "upstream" in a text-to-speech (TTS) pipeline; by directly tapping into the same intermediate audio representations used to synthesize audio, our system enables the simultaneous generation of both audio and facial motion, enabling us to spend more compute time on expressiveness and controllability of the face without incurring additional latency. 

\subsection{Audio Speech Representations and Audio LLMs}
Effective speech representations are central to both audio-driven facial animation and TTS synthesis. Early research predominantly employed handcrafted features such as Mel-Frequency Cepstral Coefficients (MFCCs) \cite{karras_audio-driven_2017}. The emergence of self-supervised learning (SSL) introduced more robust alternatives, including HuBERT \cite{hsu2021hubert}, wav2vec 2.0 \cite{baevski2020wav2vec}, and WavLM \cite{chen2022wavlm}. These SSL-based models provide superior phoneme-to-viseme mappings compared to traditional methods, yielding higher accuracy in  facial animations \cite{faceformer2022}.

In TTS, the paradigm has shifted from continuous feature regression toward discrete neural tokenization, driven by advances in Audio Large Language Models (Audio-LLM). Architectures such as \cite{zeghidour2021soundstream, defossez2022high} employ Vector Quantization (VQ) or Residual Vector Quantization (RVQ) to learn compact, discrete latent representations of speech. GLM-4-Voice \cite{zeng2024glm4voice} introduces multi-objective training that jointly optimizes for ASR and audio reconstruction, producing tokens that are both acoustically faithful and semantically meaningful for LLM reasoning. CosyVoice \cite{du2024cosyvoice} adopts a hybrid approach, augmenting discrete tokens with continuous audio features to enable finer-grained control over paralinguistic behaviors.

These discrete speech representations enable seamless integration of speech capabilities into pretrained text-based LLMs through two key steps: (1) expanding the vocabulary to include speech tokens, and (2) finetuning the model on audio token data. Architecturally, this pipeline consists of a speech tokenizer that converts continuous audio wavforms into discrete semantic tokens, the LLM itself for token prediction, and an audio decoder that reconstructs continuous speech from predicted tokens (\autoref{fig:voice_llm}). This framework has enabled audio chatbots such as Qwen3-Omni \cite{xu2025qwen3omni}, MiniOmni-2B \cite{xie2024miniomni2b}, GLM-4-Voice \cite{zeng2024glm4voice}, and SpeechGPT \cite{zhang2023speechgpt}. Our work extends this pipeline by introducing a token-to-animation model (\autoref{fig:voice_llm}) that consumes token embeddings, enabling {\em audio-visual} conversational chatbots.

\section{Method}
\label{sec:method}
We propose a token-based facial animation framework compatible with existing Audio-LLMs. Our approach takes token embeddings and a short style reference clip as input, and generates stylized facial animations that synchronize with the audio while adhering to the visual style of the reference clip. To enable generalization across different audio LLM architectures, we introduce a novel and efficient adaptation framework that bridges varying token representations. 
We now describe how our system integrates with Audio-LLM pipelines (\S\ref{sec:architecture}), the design of our style-conditioned motion decoder and adaptation layer (\S\ref{sec:decoder}), and our training strategy (\S\ref{sec:training}).

\begin{figure}[h]
    \centering
    
    \includegraphics[
    width=\linewidth,
    trim=0 0.8cm 0 0,
    clip
]{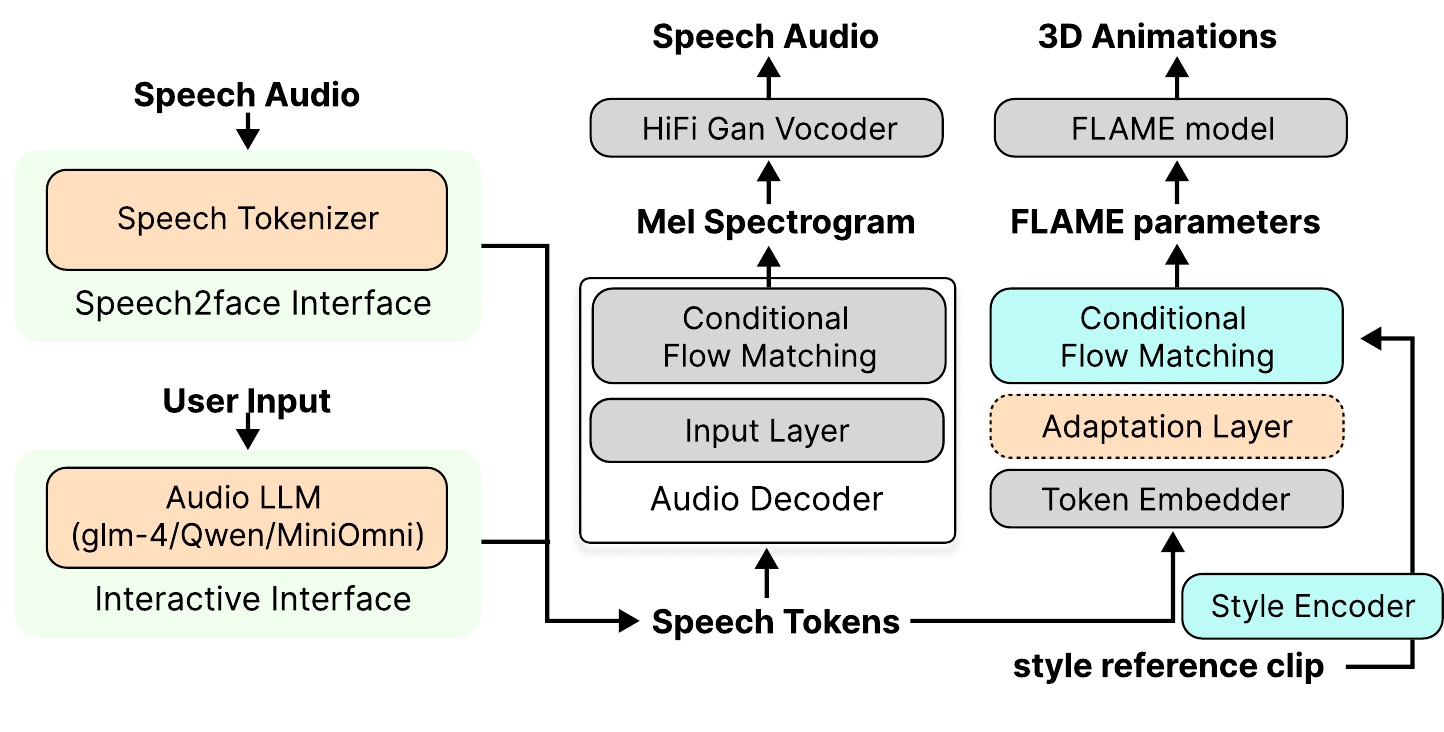}

    \caption{System Overview}
    \label{fig:sys-overview}
\end{figure}

\subsection{Pipeline Architecture}
\label{sec:architecture}
Our token-based animation model runs parallel to the audio decoder of an end-to-end Audio-LLM, acting as a decoder for motion \autoref{fig:voice_llm}. We take the token embedding as input, then use a conditional flow matching model \cite{lipman2023flow} to map the audio information to the embedding space of the FLAME model \cite{FLAME:SiggraphAsia2017}, which is then decoded into 3D vertex animation. Since different voice-LLM leverage different token spaces, we propose using a lightweight adaptation layer to map between the token spaces, ensuring our approach is generalizable to different Audio-LLM systems. 

The parallel architecture described above is enabled by our direct use of tokens embeddings, contrasting with existing speech-to-face approaches that rely on waveform-derived representations such as HuBERT \cite{hsu2021hubert} or wav2vec 2.0 \cite{baevski2020wav2vec} features. This positions our system earlier in the synthesis pipeline: while traditional methods require sequential processing (tokens → waveform → feature extraction → animation), our model consumes the same tokens as the audio decoder, enabling parallel audio-motion generation (\autoref{fig:sys-overview}). The latency implications of this design are detailed in \S\ref{sec: runtime}.
Beyond Audio-LLM integration, our token-based approach also supports conventional audio-driven facial animation. By incorporating a speech tokenizer to convert recorded audio into discrete tokens, our system can process both synthesized and recorded speech through the same token-based pipeline.

\subsection{Style Conditioned Motion Decoder} 
\paragraph{Flow Matching Model}
\label{sec:decoder}
We adopt a causal chunk-wise flow matching model with block-causal attention as our style-conditioned motion decoder. Given a sequence of audio embeddings \(e=\{e_i\}_{i=1}^{T}\), where \(e_i \in \mathbb{R}^{d_{\text{audio}}}\) from the token embedder, and a style embedding \(s\) obtained from a reference motion clip, the decoder generates FLAME-based facial mesh animations. Inspired by the audio decoder in CosyVoice~\cite{du2024cosyvoice}, our decoder operates causally over streaming chunks, allowing motion generation to proceed in parallel with audio synthesis. Formally, the conditional flow matching model \(\mathcal{G}\) learns a time-dependent vector field \(v_\theta(X_t,t,e,s)\) that transports samples from a standard normal base distribution \(p_0\) to the target FLAME motion distribution \(p_1\) through an ordinary differential equation (ODE), conditioned jointly on the audio features \(e\) and style embedding \(s\)~\cite{lipman2023flow}.

Specifically, given a FLAME parameter sequence $X_1$ where $X_1 \in \mathbb{R}^{T \times d_{\text{FLAME}}}$ represents $T$ frames of FLAME parameters (expression, jaw, and neck pose coefficients), and noise $X_0 \sim \mathcal{N}(0, I)$, we define a time-dependent interpolation path with $t \in [0, 1]$:
\begin{equation}
X_t = (1-t)X_0 + tX_1 \quad \text{where} \quad u_t(X|X_1) = \frac{dX_t}{dt} = X_1 - X_0
\end{equation}
The model is trained to match this target vector field by minimizing the conditional flow matching objective:
\begin{equation}
\mathcal{L}_{\text{CFM}}(\theta) = \mathbb{E}_{t, p(X_1), p(X_0)} \|v_\theta(X_t, t, e, s) - (X_1 - X_0)\|^2
\label{eq: flow matching}
\end{equation}
During inference, we generate FLAME parameter sequences by solving the ODE using an Euler solver:
\begin{equation}
\frac{dX_t}{dt} = v_\theta(X_t, t, e, s), \quad X_0 \sim \mathcal{N}(0, I)
\end{equation}

Using a jointly-trained style encoder $\mathcal{E}$ following MSMD~\cite{pan2025modelb}, the style embedding $s = \mathcal{E}\left(X_{\text{ref}}\right)$ is extracted from the same clip $X_{\text{ref}}$ as the target animation, and the model generates FLAME parameters $\hat{X}_{1:T} = \mathcal{G}\left(\{e_i\}_{i=1}^{T}, s\right)$.

\paragraph{Audio representation}
We extract token embeddings from the input layer of the audio decoder (\autoref{fig:sys-overview}). These representations encode both acoustic and semantic features, capturing not only phonetic content but also emotional information from speech prosody.

\paragraph{Adaptation Layer} 

To accommodate the diverse audio tokenization schemes employed by different Audio-LLMs, we propose a lightweight adaptation layer that aligns arbitrary token embedding spaces to the input space of our flow matching model. The adaptation layer is a shallow transformer that maps token embeddings from any audio decoder to match the expected input of the pre-trained flow-matching model, effectively bridging different token representations without retraining the entire system. Training occurs in two stages: we first train the flow matching model to convergence, then freeze it and train only the adaptation layer on the new Audio-LLMs. This enables efficient generalization across different Audio-LLM architectures with minimal computational overhead. Further details are provided in \S\ref{sec:training}.

\subsection{Training Strategy}
\label{sec:training}

The training consists of two stages: (1) initial training of the base flow-matching model, and (2) adaptation training. Conditioning signals are randomly dropped during training for classifier-free guidance.
The initial training starts by using the audio tokenizer and decoder to generate token embedding for the training corpus. 
During the initial training, the adaptation layer is skipped, and the base model $\mathcal{G}$ is trained through minimizing $\mathcal{L}_{\text{CFM}}$. This first stage enables our framework to work end-to-end with one specific Audio-LLM. 

We further extend our model to alternative Audio-LLMs in the second stage. This is achieved by training the adaptation layer $\mathcal{D}$ while keeping the base flow-matching model $\mathcal{G}$ frozen. For each target Audio-LLM, we construct a training corpus by extracting token embeddings $\{\tilde{e}_i\}_{i=1}^{\tilde{T}}$ from its respective tokenizer or decoder, temporarily interpolating them to match the sampling rate of the base tokenizer. To robustly project these sequence embeddings into the base embedding space, we optimize the adaptation layer using a dual objective. This comprises a feature-level alignment loss and the original flow-matching loss $\mathcal{L}_{\text{CFM}}$ with loss weight $\lambda=0.5$ 
\begin{equation}
\mathcal{L}_{\text{adapt}} = \mathcal{L}_{\text{CFM}} + \lambda \mathcal{L}_{\text{align}}, \quad \mathcal{L}_{\text{align}} = \left\|\mathcal{D}\left(\{\tilde{e}_i\}_{i=1}^{\tilde{T}}\right) - \{e_i\}_{i=1}^{T}\right\|^2
\end{equation}

\subsection{Inference}

During inference, the system processes streaming input , either generated by an Audio-LLM or extracted from audio using a speech tokenizer. Tokens arrive in sequential chunks, each spanning $C$ seconds of audio. The adaptation layer maps incoming token embeddings to the input space of the flow-matching model; this step is skipped when the embeddings already match the base token space.

The flow-matching model generates FLAME parameters causally, chunk by chunk, using overlap-conditioned temporal inpainting during ODE solving. For each chunk, the unknown region is initialized from noise and updated by the predicted velocity field, while the overlap region from the previous chunk is treated as known and overwritten at each step. A binary inpainting mask preserves the known frames and updates only the new region, maintaining temporal coherence across chunk boundaries.

We further apply classifier-free guidance~\cite{ho2022classifierfree} by modifying the predicted velocity field, with guidance weight  $w$.
\begin{equation}
v_\theta^{\text{cfg}}(X_t, t, e, s) = (1 + w) v_\theta(X_t, t, e, s)  - w\ v_\theta(X_t, t, \varnothing, \varnothing)
\end{equation}

The decoder supports different chunk sizes and ODE solver steps, enabling a trade-off between latency and generation quality, as analyzed in \autoref{sec:experiments}.
\section{Experiments}
\label{sec:experiments}

We evaluate \ours{} along three axes. First, we study whether Audio-LLM token embeddings provide a stronger conditioning signal for expressive facial animation than conventional ASR-oriented audio representations. This analysis motivates our token-based design and is conducted on RAVDESS, where emotional variation is pronounced. Second, we evaluate animation quality on the HDTF-TFHP benchmark, comparing \ours{} against prior audio-driven facial animation methods and analyzing the effects of style conditioning and cross-token adaptation. Third, we evaluate runtime and latency to verify that \ours{} is suitable for real-time Audio-LLM applications, and analyze the quality--speed trade-off induced by chunk size and flow-matching steps. 

 We evaluate all models using Mean Square Error (MSE) and Lip Vertex Error (LVE) to assess lip-sync accuracy, and Facial Dynamic Difference (FDD) and Mouth Opening Distance (MOD) ~\cite{sun_diffposetalk_2023} to measure adherence to reference style. Unless otherwise specified, our method is evaluated in the real-time setting with chunk size $C=0.4$s and $n=20$ flow-matching steps. This configuration is justified by the runtime-quality analysis in \autoref{sec: runtime-squality tradeoff}.

\subsection{Implementation}

We train and evaluate our system on the same corpus as DiffPoseTalk, combining HDTF~\cite{zhang2021flow} and TFHP. The dataset contains $1{,}052$ videos from $588$ subjects, totaling approximately $26.5$ hours of footage at $25$ fps. Following DiffPoseTalk, we split the data by speaker into $460$ training, $64$ validation, and $64$ testing subjects.

Our flow-matching decoder adopts a 1D U-Net-style Transformer architecture adapted from the conditional decoder in CosyVoice~\cite{du2024cosyvoice}. It contains two downsampling blocks, two upsampling blocks, and $12$ middle blocks. Each stage has $4$ Transformer blocks with $8$ attention heads. The adaptation layer projects source embeddings with a 1D convolution, adds positional embeddings and processes the sequence with a $4$-layer Transformer encoder.

The base flow-matching model $\mathcal{G}$ and the adaptation layer $\mathcal{D}$ are both trained on a single NVIDIA L40S using Adam with learning rate $1\times10^{-4}$. The base model is trained for $144$ hours with the conditional flow-matching objective in \autoref{eq: flow matching}; with guidance weight $w=0.3$ used at inference time. For adaptation, we freeze $\mathcal{G}$ and train only $\mathcal{D}$ for $24$ hours using the objective in \S\ref{sec:training}. 

We use Mimi~\cite{defossez2024moshi} as the base tokenizer and adapt the model to CosyVoice2~\cite{du2024cosyvoice} and GLM-4-Voice~\cite{zeng2024glm4voice} token embeddings. CosyVoice2 is used by Mini-Omni~2~\cite{xie2024miniomni2b} and LLaMA-Omni~\cite{fang2024llama}, while Mimi underlies Qwen3-Omni~\cite{xu2025qwen3omni} and Moshi~\cite{defossez2024moshi}. This lightweight adaptation enables different Audio-LLM backends to be swapped without retraining the full motion decoder.

\subsection{Audio Representation Analysis}
\label{sec:audio_representation_analysis}

We first evaluate whether Audio-LLM token embeddings provide a stronger conditioning signal for expressive facial animation than conventional ASR-oriented audio features. To isolate the effect of the audio representation, we train and evaluate all variants on RAVDESS, where speech contains clear emotional variation. We compare Audio-LLM token embeddings from GLM-4-Voice, CosyVoice2, and Moshi against ASR-based representations from wav2vec 2.0, HuBERT, and Whisper. We also include EmoTalk~\cite{peng2023emotalk} as an emotion-aware baseline that relies on explicit tag conditioning.

As shown in \autoref{tab:ravdess_emotion}, Audio-LLM token embeddings achieve competitive performance relative to conventional ASR-oriented representations. They are not uniformly best across all metrics: for example, wav2vec 2.0 obtains the lowest FDD, indicating that ASR-style features can still capture certain temporal motion dynamics effectively. However, Audio-LLM token embeddings perform strongly on reconstruction and mouth-related metrics, with CosyVoice2 achieving the lowest MSE, LVE, and MOD among all compared representations. This suggests that speech-generation tokens preserve information useful for expressive facial animation, including linguistic, acoustic, and prosodic cues. Compared with EmoTalk, all variants of \ours{} achieve lower errors despite not relying on explicit emotion or intensity tags. This indicates that the proposed token-conditioned decoder can infer expressive motion cues directly from speech representations, providing a flexible alternative to manually specified emotion conditioning.

\begin{table}[h]
    \centering
    \caption{RAVDESS evaluation of audio representations for expressive facial animation.}
    \label{tab:ravdess_emotion}
    \begin{tabular}{lcccc}
        \toprule
        \multirow{2}{*}{Model} & MSE & LVE & FDD & MOD \\
         & (mm) $\downarrow$ & (mm) $\downarrow$ & ($\times 10^{-5}$m) $\downarrow$ & (mm) $\downarrow$ \\
        \midrule
        \multicolumn{5}{l}{\textit{ASR based}} \\
        Wav2vec2 & 0.26 & 2.15 & 2.63 & 1.51 \\
        HuBERT & 0.27 & 2.26 & 3.03 & 1.65 \\
        Whisper & 0.25 & 2.07 & 2.79 & 1.41 \\
        \midrule
        \multicolumn{5}{l}{\textit{Audio LLM based}} \\
        GLM-4-Voice & 0.23 & 1.87 & 3.11 & 1.42 \\
        CosyVoice2 & 0.22 & 1.84 & 2.99 & 1.35 \\
        Moshi & 0.24 & 2.03 & 2.90 & 1.40 \\
        \midrule
        \multicolumn{5}{l}{\textit{Other baselines}} \\
        EmoTalk & 0.35 & 3.11 & 4.33 & 1.65 \\
        \bottomrule
    \end{tabular}
\end{table}

\subsection{Evaluating Generation Quality}
\label{sec: quantitative comparisona}
\paragraph{Quantitative Comparison}

We compare against three representative prior methods: DiffPoseTalk~\cite{sun_diffposetalk_2023}, MSMD~\cite{pan2025modelb}, and FaceFormer\cite{faceformer2022}. For MSMD and FaceFormer, we retrain the models on HDTF+TFHP training split using the same data protocol as \ours{}. For DiffPoseTalk, we use the official implementation and model weights. 

\autoref{tab:ablation} also includes ablations of \ours{} without style conditioning and adapted variants using CosyVoice2 and GLM-4-Voice token embeddings. Style conditioning consistently improves our base model, These improvements confirm that the reference clip provides useful motion-style information beyond the audio content alone. The adapted variants further remain close to the base model. This demonstrates that the adaptation layer can transfer the motion decoder to different Audio-LLM token spaces with only minor degradation, without retraining the full model.

Importantly, all results for \ours{} are reported under the real-time streaming setting, whereas the strongest diffusion baseline, DiffPoseTalk, is evaluated as an offline non-real-time model. Under this stricter deployment setting, \ours{} remains competitive with DiffPoseTalk, with comparable MSE, LVE, and MOD, while achieving slightly better FDD. This suggests that token-conditioned flow matching can preserve much of the motion fidelity of offline diffusion-based animation while enabling low-latency generation. This comparison highlights a key advantage of our formulation. Existing real-time facial animation systems often obtain low latency by compressing or distilling a stronger offline model, which can reduce animation quality and typically removes explicit style control. In contrast, \ours{} is designed for streaming from the outset: it generates motion in parallel with the Audio-LLM decoder and conditions on a reference style clip, enabling real-time performance without sacrificing controllability. 

\begin{table}[h]
    \centering
    \caption{Quantitative comparison of \ours{} variants, adaptations, and prior methods.}
    \label{tab:ablation}
    \small
    \begin{tabular}{lcccc}
        \toprule
        \multirow{2}{*}{Model} & MSE & LVE & FDD & MOD \\
         & (mm) $\downarrow$ & (mm) $\downarrow$ & ($\times 10^{-5}$m) $\downarrow$ & (mm) $\downarrow$ \\
        \midrule
        
        \ours{} w/o style & 2.11 & 20.73 & 4.45 & 3.15 \\
        \ours{} w/ style & 1.39 & 14.53 & 4.29 & 2.82 \\

        \midrule
        \multicolumn{5}{l}{\textit{Adaptation Layers}} \\
        CosyVoice2  & 1.37 & 14.26 & 4.34 & 2.94 \\
        GLM-4-Voice & 1.35 & 14.02 & 4.44 & 2.96 \\
        \midrule
        \multicolumn{5}{l}{\textit{Prior methods}} \\
        DiffPoseTalk  & 1.33 & 13.64 & 4.33 & 2.68 \\
        MSMD  & 1.54 & 16.07 & 11.19 & 2.65 \\
        FaceFormer  & 2.50 & 24.02 & 5.19 & 3.77 \\
        \bottomrule
    \end{tabular}
\end{table}

\begin{figure}
    \centering
    \includegraphics[width=0.9\linewidth]{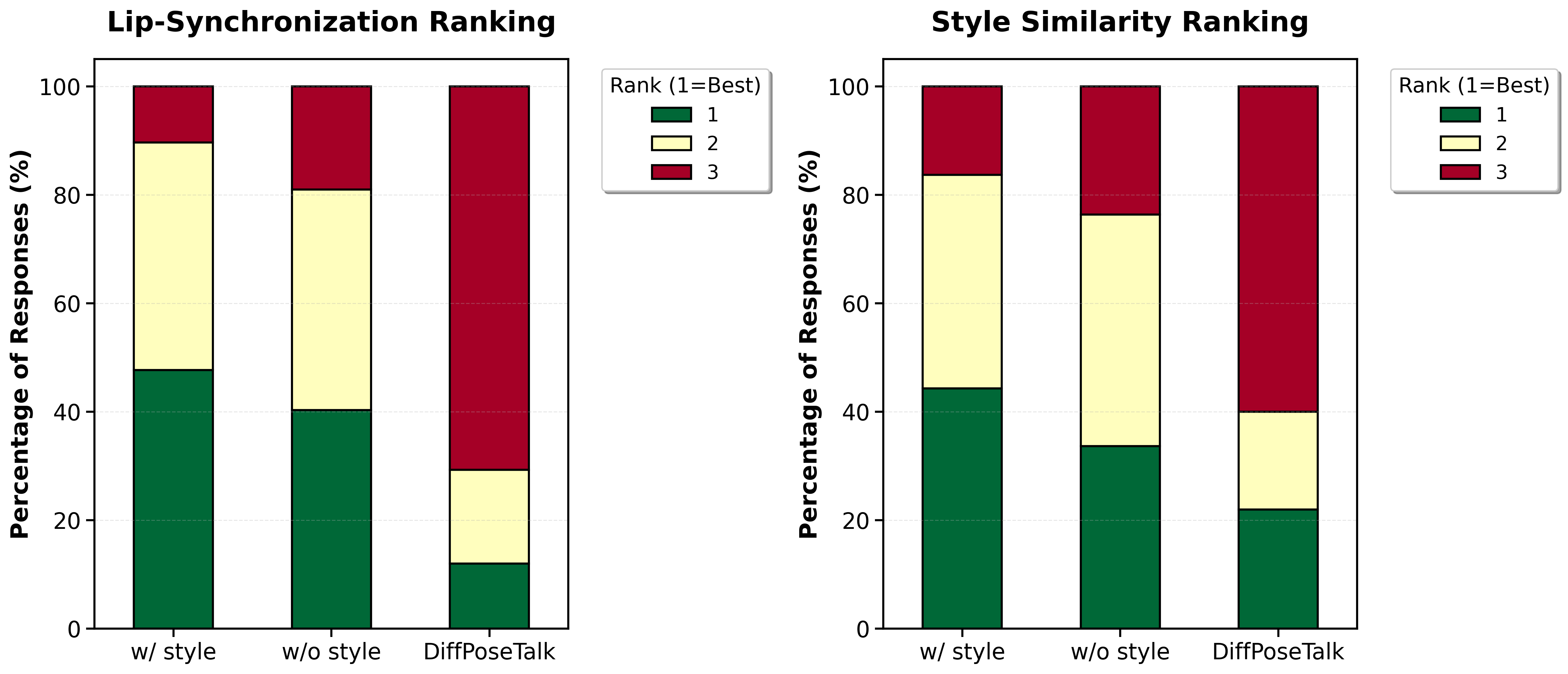}
    \caption{Perceptual ranking distributions for lip synchronization (left) and style similarity (right). Rankings are from 20 participants across 15 videos.}
    \label{fig:user study}
\end{figure}

\paragraph{Perceptual Study}
While quantitative metrics provide valuable insights, they do not fully capture perceptual quality. To complement our numerical evaluation, we conducted a user study with 15 expressive videos selected from our test set and iconic performances from YouTube. We generated animations using \ours{} (with style conditioning), \ours{} w/o style, and DiffPoseTalk~\cite{sun_diffposetalk_2023}. For each video, 20 participants viewed the ground truth alongside three generated animations in randomized order, ranking them by: (1) lip synchronization, and (2) style similarity to ground truth (300 total comparisons per metric). The results demonstrate a strong perceptual preference for our method \autoref{fig:user study}. For lip synchronization, \ours{} achieved the best mean rank, significantly outperforming the w/o style and DiffPoseTalk. A Friedman test confirmed significant differences across methods ($\chi^2 = 158.72$, $p < 10^{-34}$). For style similarity, \ours{} again ranked best (1.72), significantly outperforming both the unconditioned variant (1.90, $p < 0.01$) and DiffPoseTalk (2.38, $p < 10^{-12}$, Friedman: $\chi^2 = 69.84$, $p < 10^{-15}$).

\begin{figure*}[h]
  \centering
  \includegraphics[width=\textwidth]{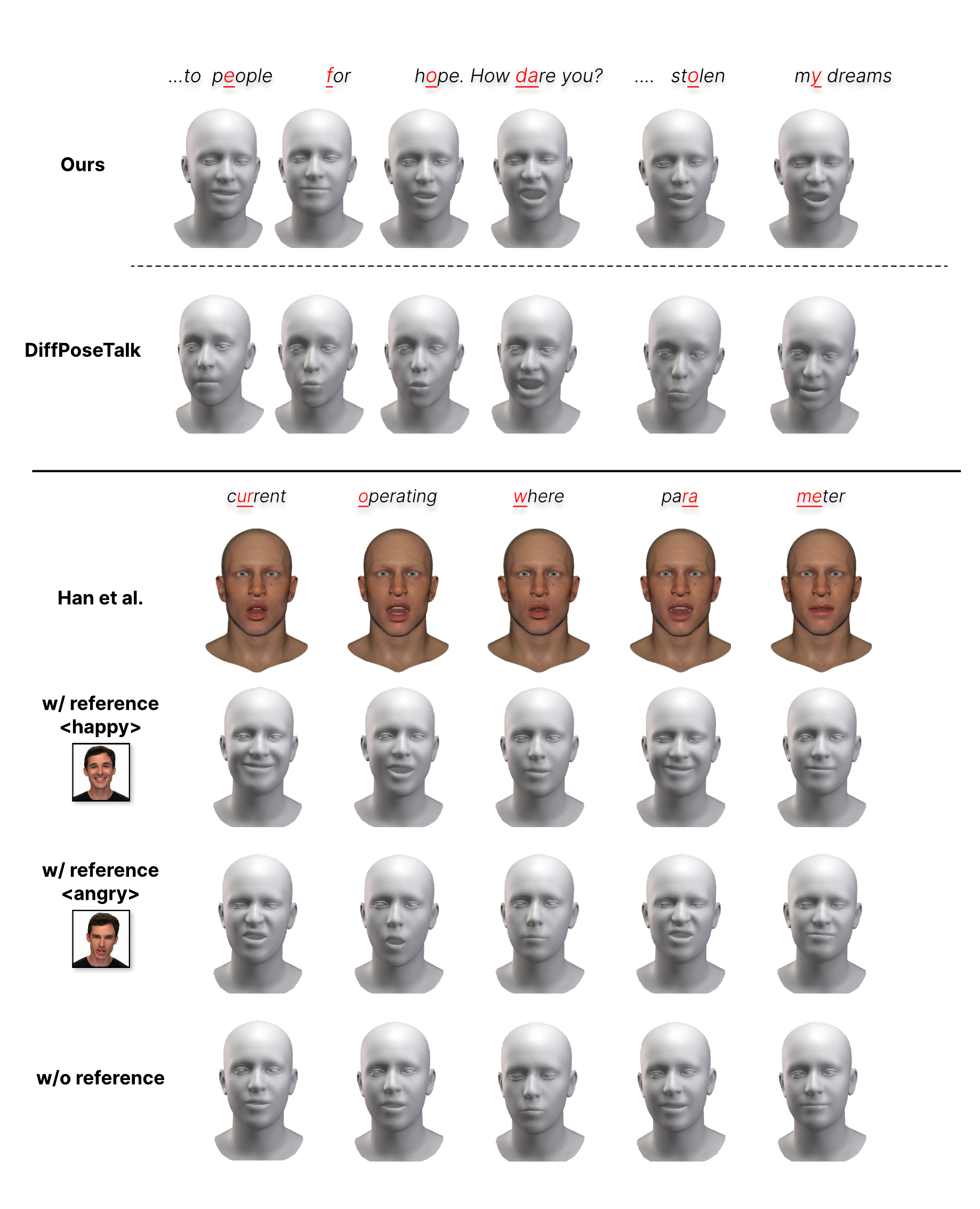}
  \caption{Qualitative comparison between \ours{}, \citet{Tiny_sig25} and \citet{sun_diffposetalk_2023}}
  \label{fig:qualitative__fig}
\end{figure*}

\paragraph{Qualitative Comparison}
We refer readers to the supplementary video for dynamic results. \autoref{fig:qualitative__fig} shows representative frames generated by \ours{} and baseline methods. Our model achieves comparable lip-sync quality to diffusion-based methods like DiffPoseTalk \cite{sun_diffposetalk_2023} while running significantly faster, and produces more dynamic and expressive animations with controllable style compared to distilled models like \citet{Tiny_sig25}.

\subsection{Runtime Evaluation}
\label{sec: runtime}
For real-time Audio LLM applications, first-frame latency (i.e. the time until the first animation frame is generated) is crucial for perceived interactivity. In a traditional cascaded pipeline using audio-driven facial animation models, the total latency to the first frame combines three components: Time to First Token Batch (TTFTB, the time for the LLM to generate enough tokens for the audio decoder to begin), audio decoding time ($T_{\text{audio}}$, inherent to the audio decoder), and animation generation time ($T_{\text{animation}}$):
\begin{equation}
    \text{Latency} = \text{TTFTB} + T_{\text{audio}} + T_{\text{animation}}
\end{equation}

However, in systems where audio decoding and animation generation occur in parallel, like \ours{}, first-frame latency becomes:
\begin{equation}
    \text{Latency} = \text{TTFTB} + \max(T_{\text{audio}}, T_{\text{animation}})
    \label{eq:latency_ours}
\end{equation}

This means that if $T_{\text{animation}} \leq T_{\text{audio}}$, animation can be generated with no additional latency cost beyond audio synthesis. In \autoref{tab:sota_comparison}, we compare the overall latency of \ours{} against state-of-the-art real-time models: \citet{Tiny_sig25} and \citet{lee2025audio}, as well as DiffPoseTalk \cite{sun_diffposetalk_2023} as a non-real-time reference. Inference times for \citet{Tiny_sig25} and \citet{lee2025audio} are taken from their original papers, while runtime measurements for DiffPoseTalk and \ours{} are computed on an Nvidia L40S GPU, averaged over 100 examples. We set $\text{TTFTB}=200$ms based on typical LLM latencies from \citet{emani2024holistic} and the minimum chunk size required by state-of-the-art audio decoders, and $T_{\text{audio}}=200$ms based on reported inference times for CosyVoice \cite{du2024cosyvoice} and GLM-4-Voice \cite{zeng2024glm4voice}.

\begin{table}[h!]
\centering
\caption{\textbf{Latency comparison with state-of-the-art baselines.} \ours{} supports variable flow matching steps ($n=1$ to $20$), yielding $T_{\text{animation}}$ ranging from 26--187ms. However, as shown in Eq.~\ref{eq:latency_ours}, overall latency remains constant since all values fall below $T_{\text{audio}}=200$ms.}
\label{tab:sota_comparison}
\resizebox{\columnwidth}{!}{
\begin{tabular}{l|c|c|c|c}
\toprule
Methods & TTFTB & $T_{\text{audio}}$ & $T_{\text{animation}}$ & Latency \\
 & (ms) & (ms) & (ms) & (ms) \\
\midrule
\citet{Tiny_sig25}  & 200 & 200 & 1.4 & 401.4 \\
\citet{lee2025audio} & 200 & 200 & 10 & 410 \\
DiffPoseTalk \cite{sun_diffposetalk_2023} & 200 & 200 & 3326 & 3726 \\
\textbf{\ours} & 200 & 200 & 26--187 & 400 \\
\bottomrule
\end{tabular}%
}
\end{table}

The results show that despite having longer $T_{\text{animation}}$ than the fastest models, \ours{} achieves the lowest latency among all methods through its efficient parallel generation scheme. Importantly, while \citet{Tiny_sig25} and \citet{lee2025audio} also achieve low latency, they do so at the cost of animation quality. \citet{Tiny_sig25} reports that its real-time variant preserves only about 70--80\% of the quality of the teacher model due to distillation, and it does not provide explicit reference-based style control, while \citet{lee2025audio}'s few-steps diffusion distillation approach is prone to introduce frequent jittering artifacts. In contrast, our model can fully leverage the available compute budget to ensure high-quality motion without incurring additional latency. We showcase a demo where we integrate our model with a voice LLM in the supplementary video.

\subsection{Runtime-Quality Trade-off}
\label{sec: runtime-squality tradeoff}
\ours's chunk-wise flow matching architecture provides flexible control over inference speed and animation quality through two key parameters: chunk size $C$ (the number of video frames generated at a time) and the number of flow matching steps $n$. Both parameters influence overall latency: chunk size affects the number of tokens required to start generation (impacting $\text{TTFTB}$), while $n$ affects $T_{\text{animation}}$. To characterize this trade-off, we evaluate \ours{} across different combinations of $C$ and $n$, measuring MSE and jitter (average per-frame acceleration near chunk boundaries), as shown in \autoref{tab:chunk_ablation_wide}.

\begin{table}[h]
    \centering
    \caption{Ablation study on chunk size ($C$) and flow matching steps ($n$). Increasing both parameters improves MSE and jitter, with stronger effects on jitter.}
    \label{tab:chunk_ablation_wide}
    
    \resizebox{\columnwidth}{!}{
    \begin{tabular}{cc cccccc}
        \toprule
        \multirow{2}{*}{\textbf{Steps}} & \multirow{2}{*}{$T_{\text{animation}}$} & \multicolumn{2}{c}{\textbf{$C = 0.4$s}} & \multicolumn{2}{c}{\textbf{$C = 0.8$s}} & \multicolumn{2}{c}{\textbf{$C = 1.2$s}} \\
        \cmidrule(lr){3-4} \cmidrule(lr){5-6} \cmidrule(lr){7-8}
        ($n$) & (ms) & MSE $\downarrow$ & Jitter $\downarrow$ & MSE $\downarrow$ & Jitter $\downarrow$ & MSE $\downarrow$ & Jitter $\downarrow$ \\
        \midrule
        1  & 26 & 1.61 & 0.232 & 1.58 & 0.246 & 1.57 & 0.239 \\
        5  & 60 & 1.57 & 0.223 & 1.55 & 0.204 & 1.54 & 0.171 \\
        10 & 101 & 1.56 & 0.201 & 1.54 & 0.184 & 1.53 & 0.151 \\
        20 & 187 & 1.57 & 0.178 & 1.53 & 0.161 & 1.52 & 0.140 \\
        \bottomrule
    \end{tabular}
    }
\end{table}

The results reveal several key insights. First, increasing both flow matching steps $n$ and chunk size $C$ consistently improves motion quality across both metrics. This informed our choice of parameters for real-time deployment: $C=0.4$s and $n=20$ ensures $T_{\text{animation}} \leq T_{\text{audio}}$ while minimizing $\text{TTFTB}$. For offline processing, larger values of both parameters would be optimal. Second, while improvements in reconstruction error are modest, gains in motion smoothness are more substantial. This echoes \citet{lee2025audio}, which demonstrated that few-step diffusion is prone to jitter. Our results confirm that maintaining boundary consistency remains a challenge for chunked diffusion-based generative models. We note that adaptation adds negligible overhead (20ms), not preventing real-time streaming.

\section{Discussion}
\subsection{Applications}
\label{sec:application}
\paragraph{Real-time Conversational Avatar} 
Leveraging \ours's low first-frame latency and fast inference time $T_{animation}$ (\autoref{sec: runtime}), our system enables two distinct online/interactive applications relevant for gaming, social media, digital assistants, and other interactive applications: an audio-LLM driven chatbot avatar, and a voice-driven user avatar (\autoref{fig:sys-overview}).\\ 
The audio for the voice-driven user avatar is tokenized as it streams, allowing users to embody an avatar using only their voice. Running with $C=0.4$ and $n=5$, the system buffers $400$ms of audio before generating animation with $60$ms inference time, resulting in a $460$ms initial delay. Once started, animation streams continuously in real-time with the user's voice.
For audio-LLM driven chatbot avatars, we operate in parallel with the audio decoder, directly consuming audio tokens generated by the language model. Note that the overall latency is bottle-necked by the token and audio generation of the LLM, \ours \ does not contribute to latency. 
\paragraph{Animation Director's Interface}
Integrating \ours{} with a voice LLM creates a director-actor paradigm for full-duplex performance generation. Given only a text script, the system generates complete audio-visual performances that can be refined through two complementary control mechanisms: 1) prompts (text or audio instructions) that modulate the LLM's emotional tone and speech cadence, which our model interprets through the audio's paralinguistic content, and 2) style reference clips that guide animation characteristics. This enables a directorial workflow where animators shape performances by instructing the voice-LLM system much as a director would to guide an actor (\autoref{fig:director-interface}). We demonstrate this interaction in the supplementary video, where generated FLAME parameters drive a production-ready character rig.

\begin{figure}[tbp]
  \includegraphics[width=0.9\columnwidth]{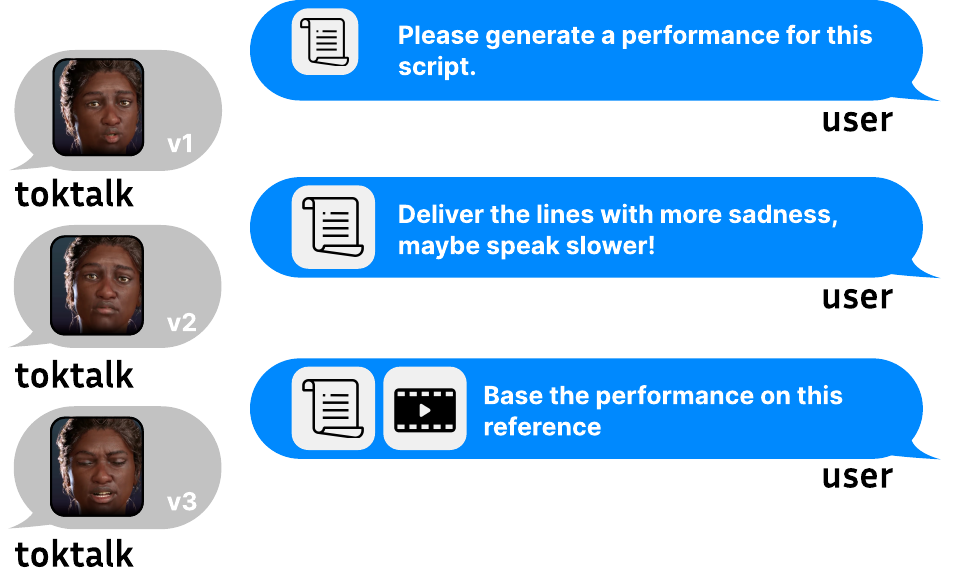}
  \caption{Multi-modal directorial interface for iterative control of audio and facial animation through text prompts and visual style references.}
  \label{fig:director-interface}
\end{figure}

\subsection{Limitations}
\label{sec:Limitations}
While our parallel structure eliminates additional latency in audio-LLM applications, speech-driven generation incurs higher latency due to our minimum chunk size of 400ms compared to autoregressive models like \cite{Tiny_sig25} (265ms lookahead). Audio tokenizer constraints further impact latency—for example, CosyVoice \cite{du2024cosyvoice} requires a minimum of 5 tokens (200ms).
Our quality-latency analysis (\autoref{sec: runtime-squality tradeoff}) reveals an inherent limitation: like \cite{lee2025audio}, fewer flow matching steps introduce jitter. While using 20 steps significantly reduces artifacts, future improvements in audio decoder speed may force our model to use fewer steps to maintain $T_{\text{animation}} \leq T_{\text{audio}}$, potentially trading quality for latency. 

\subsection{Future Work}
While our model provides facial animation for audio-LLMs with no latency cost, interactive systems must also address turn-taking detection \cite{wang2024turntakingb}, which significantly impacts response times. Additionally, natural interaction requires listening behaviors \cite{ng2022learninga} and semantic non-verbal cues such as head nods and shakes \cite{ao2023gesturediffuclip} and eye gaze \cite{pan2024s3b}, which remain important directions for future interactive systems.

\subsection{Conclusion}
\label{sec:conclusions}

We introduced \ours, a real-time, flow-matching based expressive facial animation system designed to seamlessly integrate with modern audio-LLMs. Shifting from cascaded, waveform-based processing to a parallel, token-based architecture, we achieve expressive and controllable generation without sacrificing inference speed. Our approach jointly leverages semantic and paralingual information embedded within audio tokens and style references, enabling diverse control over vivid, emotionally coherent facial expressions that are perfectly synchronized with speech. We see our work as an exciting step to transforming various text-based LLM applications to engaging human-like conversations.

{
    \small
    \bibliographystyle{ieeenat_fullname}
    \bibliography{main}
}

% WARNING: do not forget to delete the supplementary pages from your submission 
% \input{sec/X_suppl}

\end{document}